%% file: main.tex
\documentclass[letterpaper, 10 pt, conference]{ieeeconf}
\IEEEoverridecommandlockouts
\def\RALRequireImages{}
\input{preamble.tex}

\title{\LARGE\bfseries Sim(3)-TurboReg: Robust and Efficient Scale-Aware Registration via Tri-Scale Consistency}
\author{Zijun Zhou$^{*1}$ \quad Shaocheng Yan$^{*1}$\\
Zhenjun Zhao$^{2}$ \quad Yutong Hu$^{1}$ \quad Ke Yang$^{1}$ \quad Yingdong Gu$^{1}$ \quad Pengcheng Shi$^{1}$ \quad Jiayuan Li$^{1\dagger}$\\[2pt]
{\normalsize $^{1}$Wuhan University \quad $^{2}$University of Zaragoza}\\
{\normalsize \{zzjivan, shaochengyan, ljy\_whu\_2012\}@whu.edu.cn}%
\thanks{$^{*}$Equal contribution.}%
\thanks{$^{\dagger}$Corresponding author.}%
}
\hypersetup{
  pdftitle={Sim(3)-TurboReg: Robust and Efficient Scale-Aware Registration via Tri-Scale Consistency},
  pdfauthor={Zijun Zhou, Shaocheng Yan, Zhenjun Zhao, Yutong Hu, Ke Yang, Yingdong Gu, Pengcheng Shi, Jiayuan Li},
  pdfsubject={RA-L submission draft}
}

\begin{document}
\maketitle
\thispagestyle{empty}
\pagestyle{empty}

\begin{abstract}
Feed-forward 3D foundation models recover geometry at high fidelity, yet inference memory constraints keep them from processing long sequences natively.
Current streaming systems work around this limit with sliding-window chunking, which introduces severe monocular scale ambiguity.
Resolving this ambiguity requires Sim(3) registration that stays accurate and efficient under outlier contamination.
We introduce Sim(3)-TurboReg, which reformulates the compatibility graph construction of TurboReg by replacing absolute distance constraints with a strict Tri-Scale Consistency (TSC) criterion.
Our analysis shows that TSC suppresses both mismatching and coordinate-distortion outliers.
We integrate the resulting solver into existing streaming systems and obtain high-fidelity, kilometer-scale reconstruction for feed-forward models.
On controlled unknown-scale registration at 90\% outliers, ours matches RANSAC accuracy with $92.5\%$ lower median latency.
In DA3-Streaming on KITTI, ours improves mean ATE RMSE by $22.2\%$ and reduces mean end-to-end time by $16.6\%$.
Code: \url{https://github.com/ivan-zzj/Sim3-TurboReg}.
\end{abstract}

% Official RA-L list; 3 entries only. Reconstruction-oriented, not SLAM/Localization.
\begin{keywords}
Mapping, RGB-D Perception, Deep Learning for Visual Perception
\end{keywords}

\input{intro.tex}
\input{related.tex}
\input{method.tex}
\input{exp.tex}

\input{conclusion.tex}

\FloatBarrier
\bibliographystyle{IEEEtran}
\bibliography{main}
\end{document}

%% file: preamble.tex
\usepackage[T1]{fontenc}
\usepackage[utf8]{inputenc}
\usepackage{amsmath,amssymb,mathtools,bm,dsfont}
\usepackage{graphicx}
\usepackage[table]{xcolor}
\usepackage{booktabs,multirow,makecell,array}
\usepackage{algorithm}
\usepackage{algpseudocode}
\usepackage{pifont}
\usepackage{cite}
\usepackage{xspace}
\usepackage{microtype}
\usepackage{afterpage}
\usepackage{placeins}
\usepackage{stfloats}
\usepackage{tikz}
\usetikzlibrary{arrows.meta,positioning,calc,decorations.pathreplacing,shapes.geometric,backgrounds,fit}
\usepackage[caption=false,font=footnotesize]{subfig}
\definecolor{linkblue}{RGB}{0,102,204}
\usepackage[colorlinks,linkcolor=linkblue,citecolor=linkblue,urlcolor=linkblue]{hyperref}

\renewcommand{\footnoterule}{\kern-3pt \hrule width 0.4\columnwidth \kern 2.6pt}

\renewcommand{\baselinestretch}{1.01}
\normalsize

\let\oldthebibliography\thebibliography
\renewcommand{\thebibliography}[1]{%
  \oldthebibliography{#1}%
  \fontsize{7.0}{8.0}\selectfont%
  \setlength{\itemsep}{0pt}%
  \setlength{\parskip}{0pt}%
  \setlength{\parsep}{0pt}%
}

\newcommand{\cmark}{\textcolor{green!70!black}{\ding{51}}}
\newcommand{\xmark}{\textcolor{red}{\ding{55}}}
\newcommand{\req}{\textcolor{red}{\textit{Required}}}
\newcommand{\noneed}{\textcolor{green!60!black}{\textit{No Need}}}
\newcommand{\sparse}{\textcolor{red}{\textit{Sparse}}}
\newcommand{\dense}{\textcolor{green!60!black}{\textit{Dense}}}

\newlength{\ralimagewidth}
\newlength{\ralimageheight}
\newcommand{\ralimage}[3]{%
  \IfFileExists{#3}{%
    \includegraphics[width=#1,height=#2,keepaspectratio]{#3}%
  }{%
    \ifdefined\RALRequireImages
      \PackageError{ral-draft}{Missing image: #3}{Restore all images before submission.}%
    \else
      \PackageWarningNoLine{ral-draft}{Image placeholder: #3}%
    \fi
    \begingroup
    \setlength{\ralimagewidth}{#1}%
    \setlength{\ralimageheight}{#2}%
    \setlength{\fboxsep}{6pt}%
    \fbox{\parbox[c][\dimexpr\ralimageheight-2\fboxsep-2\fboxrule\relax][c]{\dimexpr\ralimagewidth-2\fboxsep-2\fboxrule\relax}{%
      \centering\small Image placeholder\par\medskip
      \footnotesize\nolinkurl{#3}%
    }}%
    \endgroup
  }%
}

%% file: intro.tex
\section{Introduction}
\label{sec:intro}

\begin{figure*}[t]
\centering
\includegraphics[width=1\linewidth]{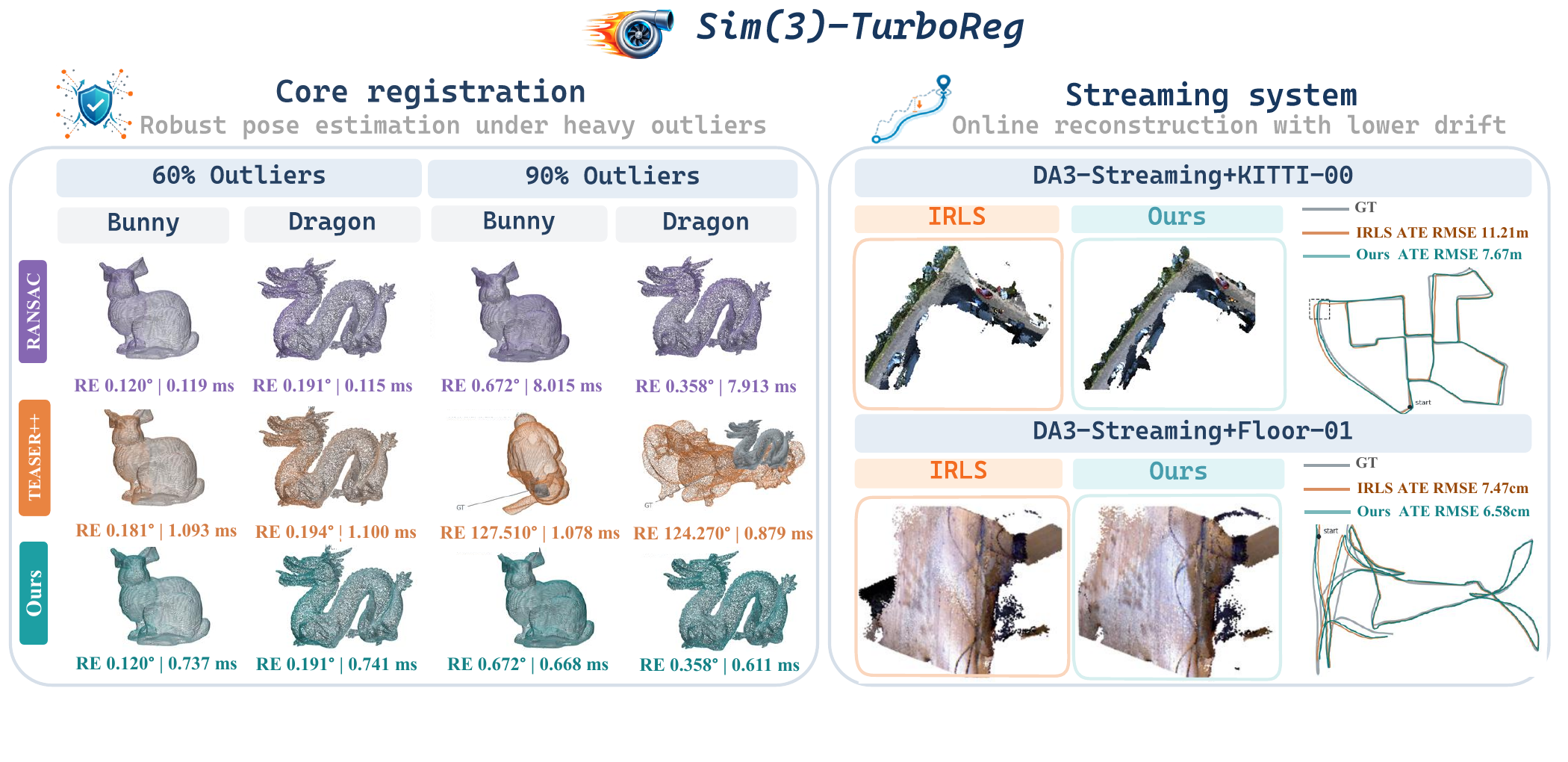}
\vspace{-2.5em}
\caption{Sim(3)-TurboReg overview. \textbf{Left:} core registration on Stanford Bunny and Dragon at 60\% and 90\% outliers. At 90\% outliers, TEASER++ fails while our method matches RANSAC accuracy at lower latency. \textbf{Right:} streaming integration with DA3-Streaming on KITTI-00 and TUM Floor-01, comparing IRLS and our backend. Sequence ATE RMSE decreases from 11.21 to 7.67\,m (31.6\%) on KITTI-00 and from 7.47 to 6.58\,cm (11.9\%) on Floor-01.}
\vspace{0.8em}
\label{fig:overview}
\end{figure*}

Feed-forward 3D foundation models~\cite{dust3r,mast3r,wang2025vggt} are now a standard approach to visual geometry, recovering dense pointmaps and camera poses from uncalibrated images in a single forward pass. However, their joint attention mechanisms impose memory costs that grow quadratically with the input length~\cite{vaswani2017attention}, causing out-of-memory failures on sequences beyond a few hundred frames on a single GPU~\cite{bertasius2021space}. This computational ceiling prevents their deployment on the continuous, kilometer-scale video streams required for autonomous driving~\cite{caesar2020nuscenes} and large-scale mapping~\cite{cadena2016past}.

The prevailing workaround for this memory bottleneck is a sliding-window strategy~\cite{droid_slam,vggt_long}: partitioning a long video into overlapping chunks, processing each independently, and aligning the resulting local reconstructions into a coherent global map~\cite{droid_slam}. 
However, without metric priors, feed-forward models reconstruct each chunk in its own local coordinate system with an unconstrained scale~\cite{dust3r, mast3r}.
This inherent \textbf{scale ambiguity} systematically invalidates standard rigid SE(3) registration, which relies on absolute distance preservation. 
Consequently, to fuse these independent chunks, the alignment task is raised from SE(3) to the more challenging Sim(3) domain~\cite{strasdat2010scale, umeyama1991least}.
Fig.~\ref{fig:overview} illustrates the resulting global reconstruction and compares local chunk alignment using IRLS and Sim(3)-TurboReg.

Robust Sim(3) solvers such as TEASER++~\cite{yang2020teaser} estimate scale, rotation, and translation despite incorrect correspondences, but accuracy and runtime depend on outlier rate and problem size. 
Conversely, sampling-based approaches (e.g., RANSAC~\cite{fischler1981random} + Umeyama~\cite{umeyama1991least}) demand excessive iterations to filter outliers, failing to guarantee deterministic solutions. 
While recent graph-based methods~\cite{turboreg2025,zhang20233dmac,chen2022sc2} achieve robustness and computational efficiency, their reliance on absolute distance invariance limits them to SE(3) registration.

To bridge this gap, we show a key insight that state-of-the-art graph-based SE(3) estimators~\cite{turboreg2025,chen2022sc2,mac} can be adapted to Sim(3) registration by changing how their compatibility graph is built.
Building on this, we introduce \textbf{Sim(3)-TurboReg}, an approach that replaces the absolute distance constraints of the original TurboReg~\cite{turboreg2025} with a strict \textbf{Tri-Scale Consistency (TSC)} criterion. 
As illustrated in Fig.~\ref{fig:pipeline}, TSC combines one edge scale and two endpoint local scales to assess correspondence compatibility.
The resulting pipeline (Fig.~\ref{fig:pipeline}) handles both general point cloud matches and the outlier distributions in streaming systems.

Unlike general point cloud registration, where feature matching techniques inevitably introduce \textbf{mismatching} outliers from incorrect associations, streaming systems that use feed-forward networks with sliding-window chunks~\cite{vggt_long} extract correspondences directly from overlapping frames after confidence filtering. 
Since these correspondences originate from identical pixels, their matching relationships are correct by construction. 
However, dynamics or depth artifacts introduce \textbf{coordinate-distortion} outliers: identities remain correct, but their coordinates no longer follow the common scene transformation.
Fig.~\ref{fig:tsc} contrasts incorrect correspondence identities with coherent coordinate distortion.
%Crucially, rather than treating these geometric outliers separately, our TSC framework provides a unified defense against both.
We then analyze why TSC suppresses mismatching outliers super-linearly by minimizing false-pass probabilities, and rejects coordinate-distortion outliers by exploiting the scale drift caused by mixed dynamic-static neighborhoods.

Our contributions are as follows:
\begin{enumerate}
    \item We propose Sim(3)-TurboReg, establishing a robust and efficient framework for Sim(3) registration by reformulating the compatibility graph construction of the original TurboReg with a strict TSC criterion.
    \item We provide a theoretical analysis showing that TSC is effective against both mismatching and coordinate-distortion outliers.
    \item We achieve a better accuracy--efficiency trade-off than prior robust Sim(3) solvers on unknown-scale registration, and as a drop-in backend improve the accuracy--runtime trade-off of streaming systems.
\end{enumerate}

% The wide pipeline figure is declared at the start of method.tex so that it
% lands at the top of the page where the Method section opens.

%% file: related.tex
\section{Related Work}
\label{sec:related}

\subsection{Feed-Forward 3D Foundation Models}
Recent advances in visual geometry are driven by feed-forward models that infer 3D structure directly from images.
On the monocular side, Depth Anything~\cite{yang2024depth} and Metric3D~\cite{yin2023metric3d} provide strong single-image geometry predictions, where Metric3D explicitly targets zero-shot metric-scale depth and surface normals across diverse camera settings.
Beyond single images, DUSt3R~\cite{dust3r} introduces dense pointmap regression for unconstrained reconstruction from an uncalibrated image pair.
MASt3R~\cite{mast3r} improves correspondence accuracy by augmenting DUSt3R with a dense local-feature head trained via an additional matching loss.
Extending this paradigm to larger view sets, VGGT~\cite{wang2025vggt} jointly predicts pointmaps and camera poses from up to hundreds of views via Transformer aggregation.
Depth Anything 3~\cite{depthanything3} extracts spatially consistent geometry from arbitrary visual inputs regardless of known camera poses.
To reduce designated reference view sensitivity, $\pi^3$~\cite{wang2025pi3} adopts a permutation-equivariant formulation.
For continuous streams, CUT3R~\cite{cut3r} updates a persistent recurrent state with each observation for online reconstruction.
Although these models provide exceptional geometry, GPU memory severely limits long continuous sequences. 
Generalizing them to streaming continuous video reconstruction remains challenging.

\subsection{Streaming and Large-Scale Reconstruction}
Classical SLAM (Simultaneous Localization and Mapping) systems like ORB-SLAM2~\cite{orbslam2}, ORB-SLAM3~\cite{orbslam3}, and LDSO~\cite{ldso} provide robust trajectory estimation. 
They typically require camera intrinsics and sensor extrinsics to build sparse maps from keypoints or high-gradient pixels.
Learning-based systems like DROID-SLAM~\cite{droid_slam}, DPVO~\cite{dpvo}, and DPV-SLAM~\cite{dpvslam} improve robustness in challenging scenarios. 
MASt3R-SLAM~\cite{mast3r_slam} integrates MASt3R~\cite{mast3r} into an online dense SLAM system featuring global optimization and loop closure. 
VGGT-Long~\cite{vggt_long} uses a chunk-and-align strategy with confidence weighting, IRLS-based~\cite{HollandWelsch1977IRLS} alignment, and a lightweight loop-closure module. 
S-MUSt3R~\cite{smust3r} employs MUSt3R-style~\cite{cabon2025must3r} priors in a similar segment-and-align pipeline adding segment-level loop detection and pose-graph optimization.
LONG3R~\cite{long3r} maintains recurrent spatio-temporal memory, while LongStream~\cite{longstream} introduces gauge-decoupled streaming geometry for extremely long sequences. 
LASER~\cite{laser} resolves cross-window scale misalignment via training-free layer-wise scale alignment. 
Depth Anything 3 (DA3)~\cite{depthanything3} provides DA3-Streaming~\cite{da3_streaming} for memory-efficient chunked inference on extended videos.
Across these systems, identity mismatches and structured geometric distortions (e.g., dynamic objects or depth artifacts) often compromise cross-window constraints. 
This difficulty necessitates robust Sim(3) alignment methods.

\subsection{Robust Registration and Similarity Estimation}
Point cloud registration estimates geometric transformations from putative correspondences corrupted by noise and outliers.
RANSAC~\cite{fischler1981random} improves robustness via minimal-set sampling, though runtime grows rapidly with low inlier ratios.
Probabilistic methods like CPD~\cite{myronenko2010point} and GMMReg~\cite{jian2010robust} formulate registration as density alignment, tolerating moderate noise but degrading under high outlier rates.
Beyond these estimators, robust pipelines explicitly prune outliers before transformation fitting.
Graph-based methods build correspondence compatibility graphs to extract self-consistent subsets.
SC$^2$-PCR~\cite{chen2022sc2} incorporates second-order spatial compatibility. 
Clique-based formulations select mutually consistent correspondences via clique discovery or graph-search strategies like MAC~\cite{mac} and FastMAC~\cite{zhang2024fastmac}.
TurboReg~\cite{turboreg2025} accelerates consistent-clique search via guided exploration.
However, these criteria primarily target SE(3) settings preserving absolute inter-point distances. 
Under global scale changes, distances scale uniformly, invalidating absolute distance checks.
In Sim(3) registration, Umeyama~\cite{umeyama1991least} provides a closed-form least-squares solution for clean correspondences.
TEASER++~\cite{yang2020teaser} improves robustness via a certifiable formulation decoupling scale, rotation, and translation under a truncated least-squares objective.

%% file: method.tex
% Declared here (not in intro.tex) so the float first becomes eligible on the
% page where the Method section opens, landing at its top.
\begin{figure*}[t]
\centering
\includegraphics[width=1\linewidth]{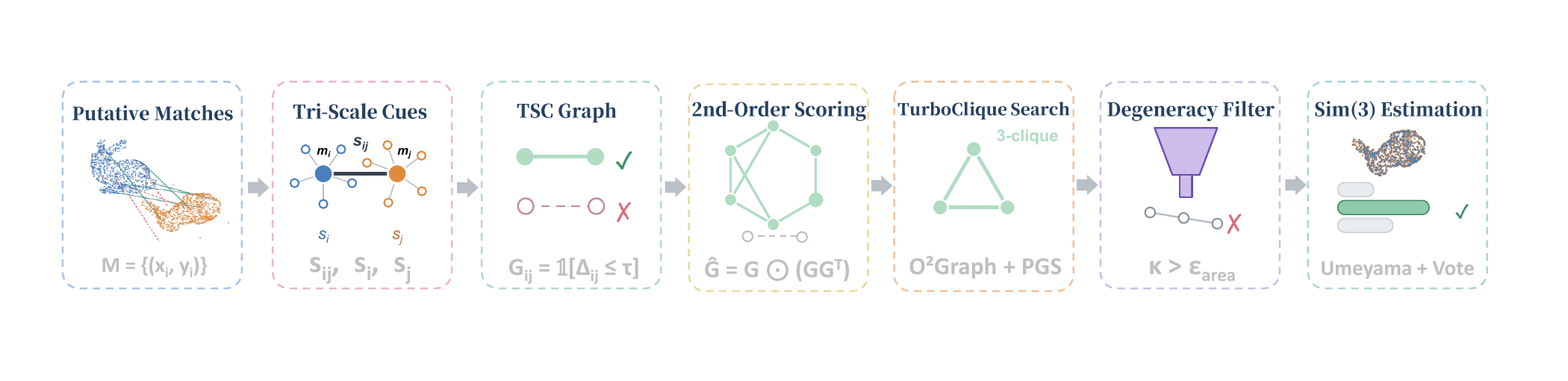}
\vspace{-3.2em}
\caption{Pipeline of Sim(3)-TurboReg. Putative matches are converted to tri-scale cues: an edge scale $s_{ij}$ and two endpoint local scales $s_i, s_j$. The TSC graph retains edges whose three scales agree ($\Delta_{ij}\le\tau$), and 2nd-order scoring weights edges by shared-neighbor support. TurboClique extracts triplets, followed by a degeneracy filter for near-collinear configurations. Umeyama hypotheses are ranked by full-set inlier voting to obtain the Sim(3) estimate.}
\vspace{0.8em}
\label{fig:pipeline}
\end{figure*}

\section{Method}
\label{sec:method}
Fig.~\ref{fig:pipeline} summarizes Sim(3)-TurboReg.
We first define the Sim(3) setting and two outlier modes in Section~\ref{sec:def_outliers}.
Sim(3)-TurboReg then proceeds in three stages:
(i) compatibility graph construction based on TSC (Section~\ref{sec:tsc}),
(ii) TurboClique solver with scale-aware degeneracy check (Section~\ref{sec:solver}),
and (iii) closed-form Sim(3) estimation with global inlier scoring (Section~\ref{sec:sim3_est}).
Finally, Section~\ref{sec:theory} explains why this design suppresses both outlier types.

% ----------------------------------------------------------

\subsection{Problem Definition and Outlier Types}
\label{sec:def_outliers}
\noindent\textbf{Setup.}\quad
Given source and target point sets $\mathcal{X},\mathcal{Y}\subseteq\mathbb{R}^3$ and putative correspondences
$\mathcal{M}=\{m_i=(\bm{x}_i,\bm{y}_i)\}_{i=1}^N$,
we estimate
$\mathbf{T}=(s,\mathbf{R},\bm{t})\in \mathrm{Sim}(3)$:
\begin{equation}
\label{eq:generative}
\bm{y}_i = s\,\mathbf{R}\,\bm{x}_i + \bm{t} + \bm{e}_i.
\end{equation}
Here, $s>0$ is the global scale, $\mathbf{R}\in SO(3)$ is rotation, $\bm{t}\in\mathbb{R}^3$ is translation, and $\bm{e}_i\in\mathbb{R}^3$ is residual noise.
For inliers, $\|\bm{e}_i\|_2\le \epsilon$.
Outliers are split into two modes:
\begin{itemize}
    \item \textbf{Type\,1 (mismatching outliers):} Correspondence identities are incorrect, as illustrated by the crossed associations in Fig.~\ref{fig:tsc}(a); we model the resulting errors $\bm{e}_i$ as unstructured noise.
    \item \textbf{Type\,2 (coordinate-distortion outliers):} Correspondence identities are correct, but coordinates are distorted by dynamics or depth artifacts. Fig.~\ref{fig:tsc}(b) shows correctly identified points undergoing coherent displacement relative to static neighbors. Unlike Type\,1, the resulting errors $\bm{e}_i$ contain \textit{structured information}, exhibiting spatial coherence or local kinematic consistency rather than random scatter.
\end{itemize}

% ----------------------------------------------------------
\subsection{Compatibility Graph Construction Based on TSC}
\label{sec:tsc}

\noindent\textbf{Edge scale ($s_{ij}$).}\quad
For a pair of correspondences $(m_i,m_j)$:
\begin{equation}
\label{eq:edge_scale}
s_{ij}=\frac{\|\bm{y}_i-\bm{y}_j\|_2}{\|\bm{x}_i-\bm{x}_j\|_2}.
\end{equation}
For true inlier pairs, $s_{ij}\approx s^*$.
If two source points are too close, the denominator in Eq.~\eqref{eq:edge_scale} becomes tiny, so small coordinate noise can excessively inflate $s_{ij}$ and destabilize estimation.
Thus, pairs with $\|\bm{x}_i-\bm{x}_j\|_2<d_{\min}$ are excluded.

\noindent\textbf{Node scale ($s_i,s_j$).}\quad
For each correspondence $m_i=(\bm{x}_i,\bm{y}_i)$, we build a bilateral neighborhood and reduce it to one robust local scale.
The two sides of $m_i$ are queried separately:
\begin{enumerate}
    \item \textbf{Source-side neighbors.} Find the $k$ nearest neighbors of $\bm{x}_i$ in $\mathcal{X}=\{\bm{x}_1,\dots,\bm{x}_N\}$, excluding the query point itself; write their correspondence indices as $\mathcal{N}_i^x$.
    \item \textbf{Target-side neighbors.} Find the $k$ nearest neighbors of $\bm{y}_i$ in $\mathcal{Y}=\{\bm{y}_1,\dots,\bm{y}_N\}$, again excluding the query point itself; write their indices as $\mathcal{N}_i^y$.
    \item \textbf{Union.} Set $\mathcal{N}_i=\mathcal{N}_i^x\cup\mathcal{N}_i^y$. The two $k$NN lists may overlap, so $|\mathcal{N}_i|$ ranges from $k$ to $2k$.
\end{enumerate}
Each index $j\in\mathcal{N}_i$ names a full correspondence $m_j=(\bm{x}_j,\bm{y}_j)$, so it contributes one \emph{pairwise scale ratio}
\begin{equation}
\label{eq:pair_ratio}
r_{ij}=\frac{\|\bm{y}_i-\bm{y}_j\|_2}{\|\bm{x}_i-\bm{x}_j\|_2}.
\end{equation}
Pairs with $\|\bm{x}_i-\bm{x}_j\|_2<d_{\min}$ are dropped (the denominator would be unstable).
The node scale is the \emph{median of these ratios}:
\begin{equation}
\label{eq:node_scale}
s_i=\operatorname{median}\bigl\{\,r_{ij}\ :\ j\in\mathcal{N}_i,\ \|\bm{x}_i-\bm{x}_j\|_2\ge d_{\min}\,\bigr\}.
\end{equation}
That is, we first form $k$--$2k$ pairwise ratios and then take their median; we do \emph{not} take a median distance first and divide.
Here, $k$ controls locality of scale consensus:
small $k$ yields high variance, while large $k$ increases robustness but may mix unrelated structure.
Using the bilateral union neighborhood avoids reliance on one coordinate space and improves robustness to structured distortion.
Algorithm~\ref{alg:tsc} summarizes the full procedure, with node-scale estimation highlighted as the key step.

\begin{algorithm}[t]
\caption{TSC Compatibility Graph Construction}
\label{alg:tsc}
\begin{algorithmic}[1]
\Require $\mathcal{M}=\{(\bm{x}_i,\bm{y}_i)\}_{i=1}^N$;\ $k$;\ $d_{\min}$;\ $\tau$
\Ensure compatibility graph $\mathbf{G}\in\{0,1\}^{N\times N}$
\For{$i=1,\dots,N$} \Comment{node scales}
  \State $\mathcal{N}_i^x\gets$ indices of $k$ nearest neighbors of $\bm{x}_i$ in $\mathcal{X}$, excluding $i$
  \State $\mathcal{N}_i^y\gets$ indices of $k$ nearest neighbors of $\bm{y}_i$ in $\mathcal{Y}$, excluding $i$
  \State $\mathcal{N}_i\gets\mathcal{N}_i^x\cup\mathcal{N}_i^y$ \Comment{$k$--$2k$ indices}
  \State $R_i\gets\emptyset$
  \For{each $j\in\mathcal{N}_i$}
    \If{$\|\bm{x}_i-\bm{x}_j\|_2\ge d_{\min}$}
      \State $R_i\gets R_i\cup\bigl\{\|\bm{y}_i-\bm{y}_j\|_2\,/\,\|\bm{x}_i-\bm{x}_j\|_2\bigr\}$ \Comment{pairwise ratio $r_{ij}$}
    \EndIf
  \EndFor
  \State $s_i\gets\operatorname{median}(R_i)$ \Comment{median of ratios, not of distances}
  \label{alg:node_scale}
\EndFor
\For{$i=1,\dots,N$} \Comment{TSC graph}
  \For{$j=i+1,\dots,N$}
    \If{$\|\bm{x}_i-\bm{x}_j\|_2\ge d_{\min}$}
      \State $s_{ij}\gets\|\bm{y}_i-\bm{y}_j\|_2\,/\,\|\bm{x}_i-\bm{x}_j\|_2$
      \State $\Delta_{ij}\gets\max(\ln s_{ij},\ln s_i,\ln s_j)-\min(\ln s_{ij},\ln s_i,\ln s_j)$
      \State $G_{ij}\gets G_{ji}\gets\mathds{1}[\Delta_{ij}\le\tau]$
    \EndIf
  \EndFor
\EndFor
\State $G_{ii}\gets 0,\ \forall i$
\end{algorithmic}
\end{algorithm}

The small-set CPU variant in Section~\ref{sec:exp_core} uses histogram node scales: it bins valid neighborhood ratios in log space and takes the geometric mean within the most populated bin. The median-based analysis below applies to Eq.~\eqref{eq:node_scale}.

\begin{figure}[t]
\centering
\includegraphics[width=\columnwidth]{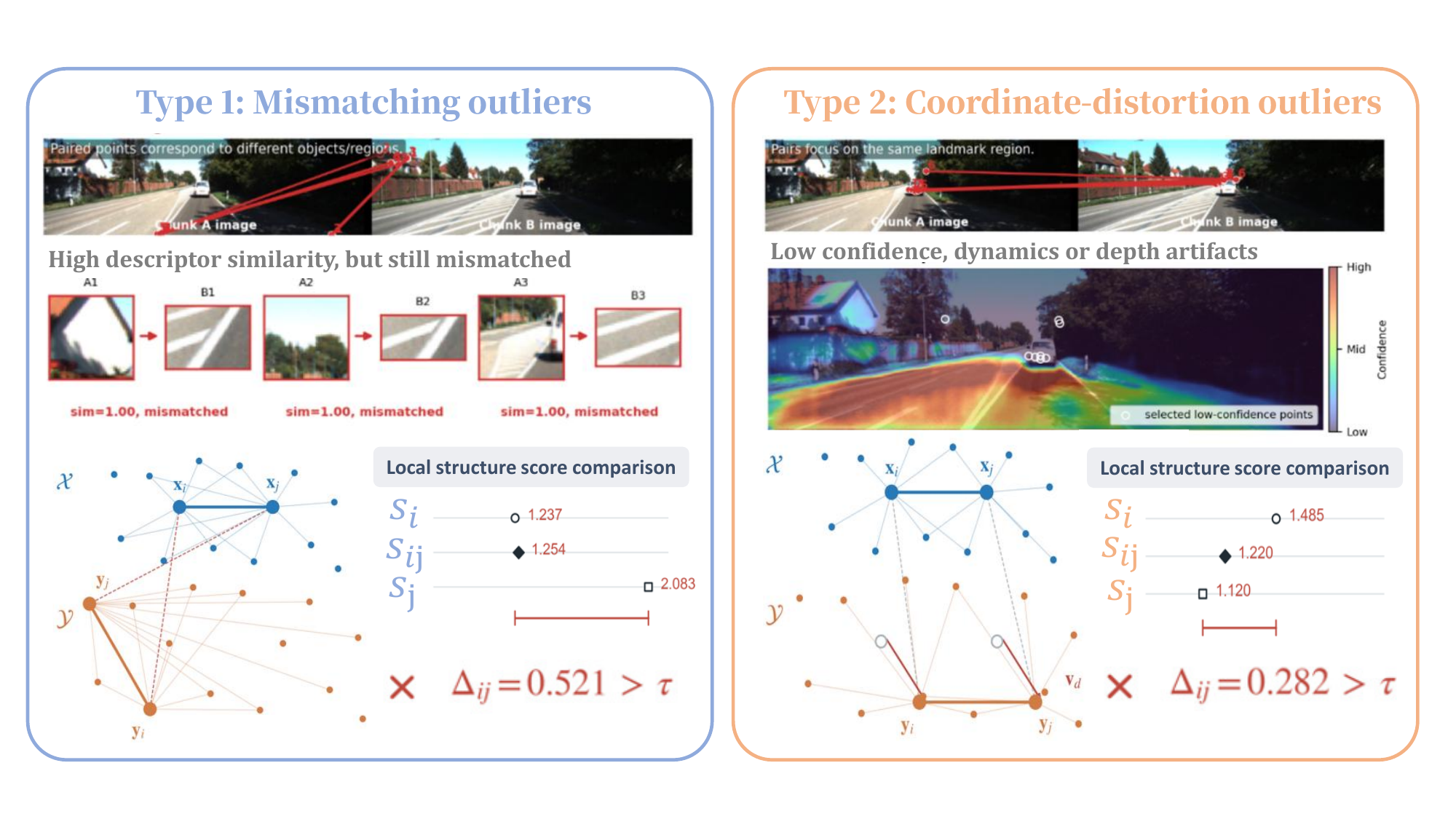}
\vspace{-1.2em}
\caption{Both outlier types in real streaming correspondences and their synthetic geometric models. (a) Type\,1 mismatching outliers: paired points link different scene regions despite high descriptor similarity. (b) Type\,2 coordinate-distortion outliers: pairs select the same landmark but undergo coherent displacement. Node scales use bilateral kNN ($k=6$), $\tau=0.15$ (illustrative settings; experiment defaults differ).}
\vspace{0.8em}
\label{fig:tsc}
\end{figure}

\noindent\textbf{TSC Graph ($\mathbf{G}$).}\quad
\label{sec:tsc_graph}
Given $(s_{ij},s_i,s_j)$, we define log-domain divergence:
\begin{equation}
\label{eq:delta}
\Delta_{ij}
=\max(\ln s_{ij},\ln s_i,\ln s_j)-\min(\ln s_{ij},\ln s_i,\ln s_j).
\end{equation}
\noindent
$\Delta_{ij}$ measures disagreement among the three scale evidences in log-space.
Consistency is
\begin{equation}
G_{ij}=\mathds{1}[\Delta_{ij}\le\tau],
\end{equation}
where $\mathds{1}[\cdot]$ is the indicator function (true $=1$, false $=0$).
We enforce $G_{ii}=0$ (no self-loops).
Smaller $\tau$ gives a purer but sparser graph, while larger $\tau$ increases recall at the cost of ambiguity.
Fig.~\ref{fig:tsc} illustrates TSC on both outlier types with real streaming correspondences and synthetic geometric examples.
The mismatched pair in Fig.~\ref{fig:tsc}(a) has $\Delta_{ij}=0.521>\tau$ due to inconsistent node scales.
In Fig.~\ref{fig:tsc}(b), a shared displacement leaves $s_{ij}=1.220$ unchanged, but the local scales become $s_i=1.485$ and $s_j=1.120$, giving $\Delta_{ij}=0.282>\tau$.
This illustrates how mixed static--dynamic neighborhoods can expose distortion even when the edge scale remains plausible; rejection depends on the neighborhood and distortion, as analyzed in Section~\ref{sec:analysis_type2}.

% Single-column TSC illustration. Declared here so the float first becomes
% eligible on page 4 and lands at its top (before the Algorithm float).

\noindent\textbf{Second-order enhancement.}\quad
Following SC$^2$-PCR~\cite{chen2022sc2}, we score mutual support:
\begin{equation}
\label{eq:sc2_tsc}
\hat{G}_{ij}=G_{ij}\sum_{u=1}^{N}G_{iu}G_{ju}.
\end{equation}
In matrix form:
\begin{equation}
\hat{\mathbf{G}}=\mathbf{G}\odot(\mathbf{G}\mathbf{G}^{\top}),
\end{equation}
where $\odot$ is the Hadamard product.
The index $u$ enumerates all correspondences and is unrelated to neighborhood size $k$.
$\hat{G}_{ij}$ is an edge score weighted by triangle support (shared neighbors).
As illustrated by the second-order scoring stage in Fig.~\ref{fig:pipeline}, edges inside an inlier-consistent subgraph receive high support, while isolated false edges remain weak.
\subsection{TurboClique Solver with Degeneracy Awareness}
\label{sec:solver}

We convert $\hat{\mathbf{G}}$ to an Ordered Graph (O$^2$Graph) and run the Pivot-Guided Search (PGS) algorithm~\cite{turboreg2025} to extract high-score 3-cliques in $O(K_1N)$ time, where $K_1$ is the number of pivots.
Unlike SE(3), Sim(3) estimates an extra scale parameter that is vulnerable to collinear or compact point configurations, requiring a strict degeneracy check for numerical stability.

\noindent\textbf{Scale-Aware Degeneracy Check.}\quad
\label{sec:degeneracy}
For the $z$-th sampled clique $\mathcal{C}_z=\{m_a,m_b,m_c\}$, we use a normalized triangle-area score:
\begin{equation}
\label{eq:kappa}
\kappa(\mathcal{C}_z)=
\frac{2\|(\bm{x}_b-\bm{x}_a)\times(\bm{x}_c-\bm{x}_a)\|_2}
{\|\bm{x}_a-\bm{x}_b\|_2^2+\|\bm{x}_b-\bm{x}_c\|_2^2+\|\bm{x}_c-\bm{x}_a\|_2^2}.
\end{equation}
$\kappa$ is scale-invariant and approaches $0$ for near-collinear triplets.
We discard cliques with $\kappa(\mathcal{C}_z)\le \epsilon_{\mathrm{area}}$.
Here, $\epsilon_{\mathrm{area}}$ is the degeneracy threshold; larger values reject more near-collinear cliques.

\subsection{Closed-Form Sim(3) Estimation}
\label{sec:sim3_est}
For each surviving clique $\mathcal{C}_z$, we estimate its corresponding local transformation parameters $T_{z}=(s_z, \mathbf{R}_z, \bm{t}_z)$ by minimizing the Umeyama objective~\cite{umeyama1991least}:
\begin{equation}
\label{eq:umeyama}
\min_{s_z,\mathbf{R}_z,\bm{t}_z}
\sum_{m_i\in\mathcal{C}_z}
\|\bm{y}_i-(s_z\mathbf{R}_z\bm{x}_i+\bm{t}_z)\|_2^2.
\end{equation}
We then perform full-set inlier voting to find the globally optimal transformation:
\begin{equation}
\label{eq:inlier_count}
\mathbf{T}
=\arg\max_{\mathbf{T}_z}
\sum_{i=1}^{N}
\mathds{1}\!\left[\|\bm{y}_i-(s_z\mathbf{R}_z\bm{x}_i+\bm{t}_z)\|_2<\tau_{\mathrm{in}}\right].
\end{equation}
Here, $\tau_{\mathrm{in}}$ is the global inlier threshold for final hypothesis scoring.

% ----------------------------------------------------------
\subsection{Analysis of TSC on Both Outlier Types}
\label{sec:theory}

\noindent
We analyze how TSC suppresses both outlier types through a two-stage mechanism: tri-scale consistency first removes most false edges, and second-order triangle support then enlarges the score gap between inlier and residual false edges.
The examples in Fig.~\ref{fig:tsc} illustrate the first stage; Fig.~\ref{fig:pipeline} depicts the subsequent shared-neighbor scoring.

\subsubsection{Type\,1: Mismatching Outliers}
\label{sec:analysis_type1}

Write $z_{ij}=\ln s_{ij}$ and $z^*=\ln s^*$.
Edge-scale-only checking (ESC) retains a false edge whenever $|z_{ij}-z^*|\le\tau$.
If the local density of Type\,1 outlier log-scales around $z^*$ is $f_0$, then
\begin{equation}
\label{eq:p_esc}
p_{\mathrm{esc}}\approx 2\tau f_0,
\end{equation}
so a mismatched edge passes as soon as its edge scale happens to be plausible.
TSC is more selective: it keeps $(i,j)$ only if all three log-scales agree with $z^*$,
\begin{equation}
\label{eq:tsc_events}
\begin{gathered}
E_{ij}\!:|z_{ij}-z^*|\le\tau,\quad
N_i\!:|\ln s_i-z^*|\le\tau,\\
N_j\!:|\ln s_j-z^*|\le\tau
\end{gathered}
\end{equation}
so $p_{\mathrm{tsc}}=P(E_{ij}\cap N_i\cap N_j)$.
For random mismatches the three evidences are largely independent; first-order factorization then gives $p_{\mathrm{tsc}}\approx p_E p_N^2$, and with $p_E\approx p_N\approx p_{\mathrm{esc}}$,
\begin{equation}
\label{eq:p_tsc}
p_{\mathrm{tsc}}\approx p_{\mathrm{esc}}^{\,3}.
\end{equation}
Correlation weakens the exponent (roughly between quadratic and cubic), but the suppression remains super-linear in the low-$p$ regime.

Second-order scoring amplifies this separation.
Let $\alpha$ be the inlier ratio, $p_+=P(G_{ij}=1\mid\text{in--in})$ and $p_-=P(G_{ij}=1\mid\text{in--out})$.
Under an SC$^2$-style count~\cite{chen2022sc2} of Eq.~\eqref{eq:sc2_tsc},
\begin{align}
\mathbb{E}[\hat G_{\mathrm{in,in}}] &\approx
p_+\!\left[(N\alpha-2)p_+^2 + N(1-\alpha)p_-^2\right],\label{eq:score_inin}\\
\mathbb{E}[\hat G_{\mathrm{in,out}}] &\approx
p_-\!\left[(N\alpha-1)p_+p_- + N(1-\alpha)p_-^2\right].\label{eq:score_inout}
\end{align}
TSC mainly reduces $p_-$ while keeping $p_+$ stable, so the gap
$\mathbb{E}[\hat G_{\mathrm{in,in}}]-\mathbb{E}[\hat G_{\mathrm{in,out}}]$ grows and TurboClique remains reliable at low inlier ratios.

\subsubsection{Type\,2: Coordinate-Distortion Outliers}
\label{sec:analysis_type2}

Type\,2 outliers are harder because two points on the same dynamic part can maintain nearly correct pairwise scale $s_{ab}\approx s^*$, so edge-only checks may retain them.
TSC defends through node scales.
Consider a mixed pair $(a,b)$ in which $a$ is displaced by $\bm{v}_d$ while $b$ stays near-static, and write $r_{ab}=\|\bm{x}_a-\bm{x}_b\|_2$, $\eta_{ab}=\|\bm{v}_d\|_2/(s^*r_{ab})$. The pairwise scale then satisfies
\begin{equation}
\label{eq:type2_exact}
\frac{s_{ab}}{s^*}
=\sqrt{1+2\eta_{ab}\cos\theta_{ab}+\eta_{ab}^2},
\end{equation}
whose log-scale perturbation is
\begin{equation}
\label{eq:type2_delta}
\delta_{ab}
=\left|\ln\frac{s_{ab}}{s^*}\right|
\approx \frac{\|\bm{v}_d\|_2}{s^*\,r_{ab}}\,|\cos\theta_{ab}|
\qquad (\eta_{ab}\ll 1).
\end{equation}
The perturbation thus grows with the displacement and shrinks with the local radius, so mixed neighborhoods with large motion or tight clustering are most sensitive.
If a fraction $\beta_i$ of node $i$'s neighbors are mixed, with typical displacement magnitude $\bar\delta_i$, the node scale drifts as
\begin{equation}
\label{eq:type2_drift}
|\ln s_i-\ln s^*|\approx c_i\,\beta_i\bar{\delta}_i,\qquad c_i\in(0,1],
\end{equation}
where $c_i$ absorbs median robustness and local geometry.
Since the dynamic-pair edge scale stays plausible ($\ln s_{ab}\approx\ln s^*$), the TSC divergence is dominated by these node-scale biases, yielding the sufficient rejection condition
\begin{equation}
\label{eq:type2_reject}
c_a\beta_a\bar{\delta}_a>\tau
\quad\text{or}\quad
c_b\beta_b\bar{\delta}_b>\tau.
\end{equation}
Type\,2 suppression therefore strengthens when distortion is larger, the mixed-neighbor ratio is higher, or the local radius is smaller.
For ray-aligned depth noise, replace $\|\bm{v}_d\|_2$ with the local depth-distortion magnitude $\sigma_{\mathrm{depth}}$; the same condition applies.

%% file: exp.tex
\section{Experiments}
\label{sec:experiments}

We evaluate core registration (Section~\ref{sec:exp_core}) and streaming integration (Section~\ref{sec:exp_system}); streaming experiments use an Intel i7-13700KF CPU and an NVIDIA RTX 4090 GPU.

\input{exp_core.tex}

\subsection{System-Level Streaming Evaluation}
\label{sec:exp_system}

We replace only the cross-chunk alignment backend of VGGT-Long~\cite{vggt_long}, Pi-Long~\cite{vggt_long,wang2025pi3}, and DA3-Streaming~\cite{depthanything3,da3_streaming}, leaving each frontend and loop-closure module unchanged.
All three align adjacent chunks with confidence-weighted IRLS under Sim(3); Pi-Long builds on $\pi^3$ rather than VGGT, and DA3-Streaming uses Depth Anything 3 in memory-efficient chunked mode.
Evaluation uses KITTI Odometry~\cite{geiger2012we} and TUM RGB-D~\cite{sturm12iros} with ATE RMSE and end-to-end runtime.

\noindent\textbf{KITTI (Table~\ref{tab:system_kitti_full}).}
DA3-Streaming improves the most and attains the best mean ATE and Avg.$^*$ among streaming pipelines; Pi-Long improves in mean ATE, though its Avg.$^*$ slightly increases (15.28$\to$15.43\,m).
The gain tracks frontend correspondence quality: dense, high-confidence matches give TSC clean neighborhoods for node scales, so the backend corrects biased chunk scales that confidence-weighted IRLS absorbs instead of rejecting.
Seq.\,01 is high-speed and its ATE is anomalously large for every streaming method, so we exclude it in Avg$^*$; aggressive viewpoint change weakens correspondence confidence and loop closure alike.
VGGT-Long is the one accuracy regression, concentrated on long sequences without effective loop closure, where biased cross-chunk estimates accumulate; its lightweight loop closure cannot absorb a biased Sim(3) chain the way Pi-Long/DA3-Streaming's denser matches can.

\input{tables/system_tum.tex}
\input{tables/system_runtime_times.tex}

\noindent\textbf{TUM RGB-D (Table~\ref{tab:tum_rgbd}).}
All three pipelines improve on average, and DA3-Streaming with our backend attains the lowest ATE on most sequences among streaming pipelines.
Indoor dynamics and depth artifacts match the Type\,2 contamination TSC targets, explaining the more uniform gain than on KITTI.

\noindent\textbf{Runtime (Table~\ref{tab:time_evaluation_turboreg}).}
Mean runtime drops in most pipeline--dataset pairings because the backend aligns chunks in milliseconds rather than seconds.
The exception is DA3-Streaming on TUM, where the plausibility gate rejects more estimates and falls back to Umeyama; cost otherwise concentrates on long KITTI routes and Pi-Long's multi-hypothesis frontend.

\noindent\textbf{Plausibility-gated fallback.}
A check (rotation $>30^\circ$ or $|\ln s|>0.3$) falls back to Umeyama on implausible estimates.
It helps VGGT-Long/Pi-Long on KITTI but slightly worsens TUM means: KITTI rejections are mostly genuine outliers, whereas on TUM some rejected estimates were correct but aggressive.

\begin{table}[!t]
\centering
\footnotesize
\setlength{\tabcolsep}{3pt}
\renewcommand{\arraystretch}{0.85}
\caption{System-level ablation (DA3-Streaming unless noted). Shaded = default. Best/second: \textbf{bold}/\underline{underlined}. KITTI Avg.$^*$ excludes Seq.\,01. Chunk rows give chunk length\,/\,overlap in frames.}
\label{tab:ablation_sim3_streaming}
\resizebox{\columnwidth}{!}{%
\begin{tabular}{@{}llcc@{}}
\toprule
& & \multicolumn{1}{c}{\textbf{KITTI (Avg.$^*$)}} & \multicolumn{1}{c@{}}{\textbf{TUM (Avg.)}} \\
\cmidrule(lr){3-3}\cmidrule(l){4-4}
& \textbf{Variant} & \textbf{ATE [m] / Time [s]} & \textbf{ATE [m] / Time [s]} \\
\midrule
\textbf{Pipe.}
& VGGT-Long~\cite{vggt_long} & 25.03 / \textbf{320} & 0.0698 / \textbf{492} \\
& Pi-Long~\cite{vggt_long,wang2025pi3} & \underline{15.43} / 1419 & \underline{0.0682} / 695 \\
& \cellcolor{gray!10}DA3-Streaming~\cite{depthanything3} & \cellcolor{gray!10}\textbf{13.76} / \cellcolor{gray!10}\underline{529} & \cellcolor{gray!10}\textbf{0.0437} / \cellcolor{gray!10}\underline{641} \\
\midrule
\textbf{Tol.}
& $\tau = 0.10$ & 17.61 / \textbf{377} & 0.0444 / 898 \\
& $\tau = 0.15$ & \underline{14.41} / 430 & 0.0448 / 938 \\
& \cellcolor{gray!10}$\tau = 0.20$ (TUM def) & 15.38 / 510 & \cellcolor{gray!10}\underline{0.0437} / \underline{641} \\
& \cellcolor{gray!10}$\tau = 0.25$ (KITTI def) & \cellcolor{gray!10}\textbf{13.76} / 529 & 0.0440 / \textbf{537} \\
& $\tau = 0.30$ & 15.42 / \underline{382} & \textbf{0.0419} / 964 \\
\midrule
\textbf{KNN}
& \cellcolor{gray!10}$k=10$ (KITTI def) & \cellcolor{gray!10}\textbf{13.76} / \cellcolor{gray!10}\underline{529} & 0.0448 / \underline{458} \\
& $k=15$ & \underline{17.97} / 637 & 0.0444 / 936 \\
& \cellcolor{gray!10}$k=20$ (TUM def) & 18.99 / \cellcolor{gray!10}\textbf{496} & \cellcolor{gray!10}\textbf{0.0437} / \cellcolor{gray!10}641 \\
& $k=25$ & 21.05 / 619 & 0.0445 / 633 \\
& $k=30$ & 18.52 / 640 & \underline{0.0440} / \textbf{456} \\
\midrule
\textbf{Inl.}
& \cellcolor{gray!10}$\tau_{\mathrm{in}} = 0.01$ (def) & \cellcolor{gray!10}\textbf{13.76} / \cellcolor{gray!10}\underline{529} & \cellcolor{gray!10}\underline{0.0437} / \cellcolor{gray!10}\underline{641} \\
& $\tau_{\mathrm{in}} = 0.02$ & \underline{15.67} / \textbf{379} & \textbf{0.0436} / \textbf{458} \\
\midrule
\textbf{Chunk}
& \cellcolor{gray!10}120\,/\,60 (def) & \cellcolor{gray!10}\underline{13.76} / \cellcolor{gray!10}\underline{529} & \cellcolor{gray!10}\textbf{0.0437} / \cellcolor{gray!10}\underline{641} \\
& 60\,/\,30 & \textbf{12.16} / \textbf{356} & \underline{0.0651} / \textbf{420} \\
\bottomrule
\end{tabular}%
}
\end{table}

\noindent\textbf{Ablation Studies.}
\label{sec:exp_ablation}
Table~\ref{tab:ablation_sim3_streaming} varies system-level choices on KITTI and TUM RGB-D.
Two patterns stand out.
First, accuracy is far more sensitive on KITTI than on TUM: across the tolerance, $k$, and $\tau_{\mathrm{in}}$ rows the TUM mean stays within $0.003$\,m, while KITTI swings by up to $7$\,m---outdoor depth ranges make local scale consensus noisier, so KITTI rewards stricter $\tau$, smaller $k$, and stricter $\tau_{\mathrm{in}}$, whereas indoor scenes saturate TSC well before these knobs matter.
Second, chunk size is the only genuine accuracy--latency trade-off: 60/30 cuts KITTI time by a third and even improves KITTI ATE, since shorter chunks drift less per alignment, but it degrades TUM, where sparser overlaps weaken the constraint chain.
The shaded defaults sit between these regimes, so $\tau$ and $k$ act as dataset-level knobs rather than per-sequence tuning.

%% file: exp_core.tex
\subsection{Controlled Unknown-Scale Registration}
\label{sec:exp_core}

\noindent\textbf{Setup.}
We evaluate on four Stanford models (Bunny, Armadillo, Dragon, Buddha)~\cite{stanford3dscanrep} following the unknown-scale protocol of TEASER~\cite[Sec.~XI-B]{yang2020teaser}.
Each trial samples $N=100$ vertices and normalizes the source cloud to unit radius, so all errors are scale-comparable across models.
Targets follow $\bm{y}_i=s^*\mathbf{R}^*\bm{x}_i+\bm{t}^*+\bm{\epsilon}_i$ with $s^*\sim\mathcal{U}[1,5]$, Haar-uniform rotation, $\bm{t}^*$ uniform in the unit ball, and Gaussian noise of scale $\sigma=0.01$ clipped to the TEASER noise bound $5.54\sigma$.
We then replace $60$--$90\%$ of the correspondences by uniform outliers drawn on the target side, i.e.\ pure Type\,1 contamination.
Two independent seeds with 40 trials per (model, rate) give $320$ trials per outlier rate and $1{,}280$ inputs per method.
We report the all-trial means of scale error $e_s=|s-s^*|$, rotation error $e_R$ (degrees), and translation error $e_t=\|\bm{t}-\bm{t}^*\|_2$, together with median single-thread CPU time (Intel i7-13700KF).
A trial is a \emph{failure} if the solver returns no estimate; the all-trial mean is then undefined and is marked $\infty$, with the failure count in parentheses.

\noindent\textbf{Baselines and implementation.}
We compare TEASER++~\cite{yang2020teaser}; RANSAC with three-point Umeyama sampling~\cite{fischler1981random,umeyama1991least}, either capped at $10^3$ iterations (\textbf{RANSAC-1K}) or run with an adaptive confidence stop under a $60$\,s budget (\textbf{RANSAC-1min}); a Sim(3) adaptation of FGR~\cite{zhou2016fast} (\textbf{FGR-Sim3}$^\dagger$, our re-implementation with a closed-form weighted similarity inside the continuation loop); and our CPU variant.
All RANSAC variants share the same $5.54\sigma$ inlier threshold and a consensus refit.
Our CPU variant uses $32$ pivots, $k=10$, histogram node scales, $d_{\min}=10^{-3}$, TSC tolerance $\tau=0.1$, and inlier-voting threshold $\tau_{\mathrm{in}}=0.08$; second-order scoring and the degeneracy filter are disabled at this scale (enabling either leaves the estimates unchanged, Table~\ref{tab:sim3_cpu_ablation}).

\input{tables/sim3_comparison_complete.tex}
\input{tables/sim3_ablation_complete.tex}

\noindent\textbf{Results.}
Table~\ref{tab:sim3_stanford} pools all models and seeds per outlier rate.
The pooled means are most informative about \emph{how} each baseline breaks.
RANSAC-1K fails catastrophically at $90\%$ outliers (80 of 320 trials): at inlier ratio $0.1$ a three-point sample is all-inlier with probability $\approx10^{-3}$, so the $10^3$-iteration cap is simply exhausted; only the adaptive variant always returns an estimate, at $13\times$ our median latency.
FGR-Sim3$^\dagger$ is the fastest but degrades earliest, because its continuation scheme reweights correspondences yet never isolates a mutually consistent subset, so once outliers dominate, the objective itself is misleading.
TEASER++ fails in a characteristic order: the scale branch degrades first ($e_s=14.3$ at $80\%$), and rotation follows at $90\%$ ($e_R=121.3^\circ$).
Decoupled scale estimation is thus the weak link under extreme contamination---exactly the quantity that TSC constrains three times per edge.
Ours is the only method besides RANSAC-1min that returns an estimate on every trial at every rate, matching its accuracy (within $0.004^\circ$ in $e_R$) at $92.5\%$ lower median latency.
Its runtime is also nearly flat in the outlier rate ($0.70$\,ms at $60\%$ vs.\ $0.61$\,ms at $90\%$): TSC rejects contamination \emph{before} search, so more outliers mean cheaper graph pruning rather than more hypotheses---the opposite of sampling-based scaling.

\noindent\textbf{Ablations.}
Table~\ref{tab:sim3_cpu_ablation} varies one factor at a time on the same $1{,}280$ inputs, and the failure pattern mirrors the method's logic.
Disabling TSC is catastrophic (910 failures): the graph becomes almost complete, so pivots land on outlier-dominated cliques whose hypotheses attract no inlier votes.
Node scales must be robust and local at once: replacing the histogram reduction with a union median occasionally lets a distorted neighborhood shift the node scale (2 failures), $k=5$ makes node scales too noisy ($e_R$ up $69\%$), and $k=20$ mixes unrelated structures until local consensus breaks (1 failure).
The risk of the tolerance is one-sided: tightening to $\tau=0.05$ costs nothing, but loosening to $\tau=0.20$ passes enough inconsistent edges to cause 5 failures---a permissive graph is more dangerous than a sparse one, because clique search trusts edge scores.
Post-refinement matters ($e_R$ $0.577^\circ$ without it), while a weighted variant adds nothing, indicating the voted inlier set is already clean.
Second-order scoring and the degeneracy filter leave every estimate unchanged at $N=100$; they matter at streaming scale (Section~\ref{sec:exp_system}), where $N$ is two orders of magnitude larger.

\afterpage{\input{tables/system_kitti.tex}}

%% file: tables/sim3_comparison_complete.tex
\begin{table}[tp]
\centering\footnotesize
\setlength{\tabcolsep}{3pt}
\renewcommand{\arraystretch}{1.0}
\caption{Unknown-scale Stanford registration (320 trials/row, single-thread CPU). Mean errors and median time; $e_t$ normalized. Errors as $e_s$ / $e_R$ ($^\circ$) / $e_t$. $\dagger$Sim(3) adaptation of FGR (our re-implementation). Best/second per metric: \textbf{bold}/\underline{underlined}. Fail.\ counts trials with no estimate (out of 320). $\infty$ marks a mean that is undefined because the method returned no estimate on at least one trial; Table~\ref{tab:sim3_cpu_ablation} uses the same convention.}
\label{tab:sim3_stanford}
\resizebox{\columnwidth}{!}{%
\begin{tabular}{@{}clrr@{}}
\toprule
Outliers & Method & $e_s$ / $e_R$ ($^\circ$) / $e_t$$\downarrow$ & CPU ms$\downarrow$ (Fail.) \\
\midrule
\multirow{5}{*}{60\%}
& RANSAC-1K & $\mathbf{0.0032}$ / $\underline{0.212}$ / $\mathbf{0.0056}$ & $\underline{0.119}$ (0) \\
& RANSAC-1min & $\mathbf{0.0032}$ / $\underline{0.212}$ / $\mathbf{0.0056}$ & $\mathbf{0.112}$ (0) \\
& FGR-Sim3$^\dagger$ & $0.1030$ / $2.052$ / $0.0994$ & $0.141$ (0) \\
& TEASER++ & $\underline{0.0061}$ / $0.261$ / $\underline{0.0076}$ & $0.914$ (0) \\
& Ours & $\mathbf{0.0032}$ / $\mathbf{0.211}$ / $\mathbf{0.0056}$ & $0.703$ (0) \\
\midrule
\multirow{5}{*}{70\%}
& RANSAC-1K & $\mathbf{0.0043}$ / $\mathbf{0.260}$ / $\mathbf{0.0071}$ & $0.260$ (0) \\
& RANSAC-1min & $\mathbf{0.0043}$ / $\mathbf{0.260}$ / $\mathbf{0.0071}$ & $\underline{0.252}$ (0) \\
& FGR-Sim3$^\dagger$ & $1.8639$ / $32.317$ / $1.6431$ & $\mathbf{0.139}$ (0) \\
& TEASER++ & $0.0679$ / $\underline{0.471}$ / $\underline{0.0363}$ & $0.891$ (0) \\
& Ours & $\underline{0.0044}$ / $\mathbf{0.260}$ / $\mathbf{0.0071}$ & $0.683$ (0) \\
\midrule
\multirow{5}{*}{80\%}
& RANSAC-1K & $\underline{0.0050}$ / $\underline{0.319}$ / $\underline{0.0086}$ & $0.875$ (0) \\
& RANSAC-1min & $\underline{0.0050}$ / $\mathbf{0.314}$ / $\mathbf{0.0085}$ & $0.860$ (0) \\
& FGR-Sim3$^\dagger$ & $5.5496$ / $73.204$ / $4.9124$ & $\mathbf{0.134}$ (0) \\
& TEASER++ & $14.3121$ / $0.702$ / $4.4651$ & $0.879$ (0) \\
& Ours & $\mathbf{0.0049}$ / $\mathbf{0.314}$ / $\mathbf{0.0085}$ & $\underline{0.671}$ (0) \\
\midrule
\multirow{5}{*}{90\%}
& RANSAC-1K & $\infty$ / $\infty$ / $\infty$ & $1.329$ (\textbf{80}) \\
& RANSAC-1min & $\mathbf{0.0074}$ / $\underline{0.487}$ / $\underline{0.0133}$ & $8.052$ (0) \\
& FGR-Sim3$^\dagger$ & $\underline{7.4818}$ / $113.084$ / $6.8162$ & $\mathbf{0.133}$ (0) \\
& TEASER++ & $41.8419$ / $121.329$ / $26.7546$ & $0.922$ (0) \\
& Ours & $\mathbf{0.0074}$ / $\mathbf{0.483}$ / $\mathbf{0.0132}$ & $\underline{0.605}$ (0) \\
\bottomrule
\end{tabular}}
\end{table}

%% file: tables/sim3_ablation_complete.tex
\begin{table}[!tb]
\centering\footnotesize
\setlength{\tabcolsep}{2.5pt}
\renewcommand{\arraystretch}{1.0}
\caption{CPU ablations on the same $1{,}280$ inputs as Table~\ref{tab:sim3_stanford}, one factor changed relative to Ours. Mean rotation error $e_R$ over all trials and median CPU time. Fail.\ counts trials with no estimate (out of $1{,}280$); a single failure makes the all-trial mean $\infty$. Ours: histogram node scales, uniform post-refinement, TSC on, $\tau=0.1$, $k=10$.}
\label{tab:sim3_cpu_ablation}
\begin{tabular*}{\columnwidth}{@{\extracolsep{\fill}}lrrr@{}}
\toprule
Change vs.\ Ours & $e_R$ ($^\circ$)$\downarrow$ & Fail.$\downarrow$ & CPU ms$\downarrow$ \\
\midrule
TSC off & $\infty$ & 910 & 0.648 \\
No post-refinement & $0.577$ & \textbf{0} & 0.620 \\
Weighted refinement & $0.317$ & \textbf{0} & 0.631 \\
Union-median node scales & $\infty$ & 2 & 0.647 \\
Second-order scoring on & $0.317$ & \textbf{0} & 0.674 \\
Degeneracy filter on & $0.317$ & \textbf{0} & 0.640 \\
$\tau=0.05$ & $0.317$ & \textbf{0} & 0.634 \\
$\tau=0.20$ & $\infty$ & 5 & 0.620 \\
$k=5$ & $0.536$ & \textbf{0} & \textbf{0.479} \\
$k=20$ & $\infty$ & 1 & 0.966 \\
\midrule
Ours (default) & $\mathbf{0.317}$ & \textbf{0} & 0.653 \\
\bottomrule
\end{tabular*}
\end{table}

%% file: tables/system_kitti.tex
\begin{table*}[!t]
\centering
\caption{KITTI ATE RMSE [m] ($\downarrow$) and mean end-to-end time [s]. $^\dagger$=prior work. Best/second among six streaming rows: \textbf{bold}/\underline{underlined}. Avg.$^*$ excludes Seq.\,01.}
\label{tab:system_kitti_full}
\resizebox{\textwidth}{!}{
\scriptsize
\setlength{\tabcolsep}{3pt}
\begin{tabular}{c | l | c c c | c | c | c | c c c c c c c c c c c}
\toprule
& \textbf{Methods} & \textbf{LC} & \textbf{Cal.} & \textbf{Recon.} & \textbf{Time [s]} & \textbf{Avg.} & \textbf{Avg.$^*$} & \textbf{00} & \textbf{01} & \textbf{02} & \textbf{03} & \textbf{04} & \textbf{05} & \textbf{06} & \textbf{07} & \textbf{08} & \textbf{09} & \textbf{10} \\
\midrule
\midrule
\multirow{3}{*}{\rotatebox{90}{\textbf{Classic}}}
& ORB-SLAM2 (w/o LC)~\cite{orbslam2}$^\dagger$& \xmark & \req & \sparse & -- & 69.727 & 26.480 & 40.65 & 502.20 & 47.82 & 0.94 & 1.30 & 29.95 & 40.82 & 16.04 & 43.09 & 38.77 & 5.42 \\
& ORB-SLAM2 (w/ LC)~\cite{orbslam2}$^\dagger$ & \cmark & \req & \sparse & -- & 54.816 & 9.464 & 6.03 & 508.34 & 14.76 & 1.02 & 1.57 & 4.04 & 11.16 & 2.19 & 38.85 & 8.39 & 6.63 \\
& LDSO~\cite{ldso}$^\dagger$              & \cmark & \req & \sparse & -- & 22.425 & 23.500 & 9.32 & 11.68 & 31.98 & 2.85 & 1.22 & 5.10 & 13.55 & 2.96 & 129.02 & 21.64 & 17.36 \\
\midrule
\multirow{8}{*}{\rotatebox{90}{\textbf{Learning Based}}}
& DROID-VO~\cite{droid_slam}$^\dagger$         & \xmark & \req & \dense  & -- & 54.188 & 51.187 & 98.43 & 84.20 & 108.80 & 2.58 & 0.93 & 59.27 & 64.40 & 24.20 & 64.55 & 71.80 & 16.91 \\
& DPV-SLAM++~\cite{dpvslam}$^\dagger$        & \cmark & \req & \sparse & -- & 25.749 & 27.138 & 8.30 & 11.86 & 39.64 & 2.50 & 0.78 & 5.74 & 11.60 & 1.52 & 110.90 & 76.70 & 13.70 \\
& VGGT-Long~\cite{vggt_long}              & \cmark & \noneed & \dense & 650 & 25.671 & 22.432 & \underline{8.63} & \textbf{58.07} & 52.89 & 8.74 & 4.22 & 9.84 & \underline{4.67} & 2.66 & 73.11 & 31.82 & 27.74 \\
& Pi-Long~\cite{vggt_long,wang2025pi3}                & \cmark & \noneed & \dense & 2337 & 29.766 & \underline{15.278} & 11.84 & 174.65 & 42.08 & 8.72 & \underline{2.42} & \underline{5.88} & 5.53 & 4.43 & 43.64 & 16.77 & \textbf{11.47} \\
& DA3-Streaming~\cite{depthanything3}     & \cmark & \noneed & \dense & 634 & \underline{24.477} & 16.541 & 11.21 & 103.84 & 71.58 & \underline{3.50} & \textbf{1.80} & \textbf{5.08} & 10.51 & \underline{2.33} & 36.19 & \underline{10.40} & \underline{12.81} \\
\rowcolor{gray!15} \cellcolor{white}
& \textbf{VGGT-Long + Sim(3)-TurboReg}      & \cmark & \noneed & \dense & \textbf{320} & 29.303 & 25.026 & 10.00 & 72.07 & \underline{41.75} & 18.55 & 3.97 & 12.30 & 4.94 & 3.20 & 97.32 & 31.62 & 26.62 \\
\rowcolor{gray!15} \cellcolor{white}
& \textbf{Pi-Long + Sim(3)-TurboReg}        & \cmark & \noneed & \dense & 1419 & 28.141 & 15.428 & 18.92 & 155.26 & \textbf{35.50} & 9.26 & 6.88 & 10.64 & \textbf{4.37} & 4.41 & \underline{33.14} & 18.28 & 12.87 \\
\rowcolor{gray!15} \cellcolor{white}
& \textbf{DA3-Streaming + Sim(3)-TurboReg}  & \cmark & \noneed & \dense & \underline{529} & \textbf{19.046} & \textbf{13.756} & \textbf{7.67} & \underline{71.95} & 47.83 & \textbf{3.34} & 3.28 & 15.04 & 9.31 & \textbf{2.08} & \textbf{24.97} & \textbf{10.20} & 13.83 \\
\bottomrule
\end{tabular}}
\end{table*}

%% file: tables/system_tum.tex
\begin{table*}[!t]
\centering
\caption{TUM RGB-D ATE RMSE [m] ($\downarrow$) and mean end-to-end time [s]. $^\dagger$=prior work. Best/second per group: \textbf{bold}/\underline{underlined}.}
\label{tab:tum_rgbd}
\resizebox{\textwidth}{!}{
\setlength{\tabcolsep}{4pt}
\begin{tabular}{c | l | c | c | c c c c c c c c c}
\toprule
\textbf{Type} & \textbf{Methods} & \textbf{Time [s]} & \textbf{Avg.} & \textbf{360} & \textbf{desk} & \textbf{desk2} & \textbf{floor} & \textbf{plant} & \textbf{room} & \textbf{rpy} & \textbf{teddy} & \textbf{xyz} \\
\midrule
\multirow{2}{*}{\shortstack{Uncalibrated\\SLAM}}
& Droid-SLAM Uncalibrated~\cite{droid_slam}$^\dagger$  & -- & \underline{0.163} & \underline{0.202} & \textbf{0.032} & \underline{0.091} & \underline{0.064} & \underline{0.045} & \underline{0.918} & \underline{0.056} & \textbf{0.045} & \textbf{0.012} \\
& Mast3r-SLAM Uncalibrated~\cite{mast3r_slam}$^\dagger$ & -- & \textbf{0.060} & \textbf{0.070} & \underline{0.035} & \textbf{0.055} & \textbf{0.056} & \textbf{0.035} & \textbf{0.118} & \textbf{0.041} & \underline{0.114} & \underline{0.020} \\
\midrule
\multirow{6}{*}{\shortstack{Streaming\\Models}}
& VGGT-Long~\cite{vggt_long}                & 868 & 0.0824 & 0.0632 & 0.0610 & 0.0601 & 0.1108 & 0.0616 & 0.1696 & 0.0374 & 0.1248 & 0.0534 \\
& Pi-Long~\cite{vggt_long,wang2025pi3}                  & 1166 & 0.0808 & 0.1254 & 0.0580 & 0.0425 & 0.0957 & 0.0660 & 0.2052 & \underline{0.0270} & \underline{0.0863} & 0.0215 \\
& DA3-Streaming~\cite{depthanything3}       & \underline{633} & \underline{0.0492} & \underline{0.0468} & \underline{0.0202} & \textbf{0.0222} & \underline{0.0747} & \underline{0.0401} & \underline{0.0948} & 0.0321 & 0.0907 & \underline{0.0208} \\
\rowcolor{gray!15} \cellcolor{white}
& \textbf{VGGT-Long + Sim(3)-TurboReg}  & \textbf{492} & 0.0698 & 0.0651 & 0.0316 & 0.0412 & 0.1203 & 0.0490 & 0.1258 & 0.0339 & 0.1168 & 0.0443 \\
\rowcolor{gray!15} \cellcolor{white}
& \textbf{Pi-Long + Sim(3)-TurboReg}  & 695 & 0.0682 & 0.0969 & 0.0295 & 0.0264 & 0.1192 & 0.0579 & 0.1403 & \textbf{0.0242} & 0.0978 & 0.0213 \\
\rowcolor{gray!15} \cellcolor{white}
& \textbf{DA3-Streaming + Sim(3)-TurboReg} & 641 & \textbf{0.0437} & \textbf{0.0459} & \textbf{0.0170} & \underline{0.0247} & \textbf{0.0658} & \textbf{0.0358} & \textbf{0.0929} & 0.0310 & \textbf{0.0615} & \textbf{0.0184} \\
\bottomrule
\end{tabular}
}
\end{table*}

%% file: tables/system_runtime_times.tex
\begin{table*}[!t]
\centering
\caption{End-to-end processing time [s] (coarse; varies with system load). Best/second: \textbf{bold}/\underline{underlined}.}
\label{tab:time_evaluation_turboreg}
\resizebox{\textwidth}{!}{
\scriptsize
\setlength{\tabcolsep}{2pt}
\begin{tabular}{l | c c c c c c c c c c c c | c c c c c c c c c c}
\toprule
\multirow{2}{*}{\textbf{Method}} & \multicolumn{12}{c|}{\textbf{KITTI Odometry}} & \multicolumn{10}{c}{\textbf{TUM RGB-D}} \\
\cmidrule{2-23}
& \textbf{Avg.} & \textbf{00} & \textbf{01} & \textbf{02} & \textbf{03} & \textbf{04} & \textbf{05} & \textbf{06} & \textbf{07} & \textbf{08} & \textbf{09} & \textbf{10} & \textbf{Avg.} & \textbf{360} & \textbf{desk} & \textbf{desk2} & \textbf{floor} & \textbf{plant} & \textbf{room} & \textbf{rpy} & \textbf{teddy} & \textbf{xyz} \\
\midrule
VGGT-Long~\cite{vggt_long}
& 650 & 1785 & 299 & 1356 & 217 & 70 & 853 & 377 & \underline{311} & \underline{1109} & 452 & 322
& 868 & 789 & 577 & 456 & 861 & 1039 & 1313 & 737 & 1238 & 800 \\
Pi-Long~\cite{vggt_long,wang2025pi3}
& 2337 & 5651 & 761 & 4481 & 649 & 263 & 3621 & 1261 & 1171 & 4525 & 2154 & 1168
& 1166 & 932 & 531 & 643 & 1433 & 1093 & 1759 & 1110 & 1852 & 1144  \\
DA3-Streaming~\cite{depthanything3}
& 634 & 900 & 233 & 1597 & 228 & 59 & 869 & 490 & 474 & 1391 & 413 & 324
& \underline{633} & 740 & \underline{330} & 380 & \textbf{704} & \textbf{719} & \underline{762} & 491 & 1096 & 472   \\
\midrule
\rowcolor{gray!15}
\textbf{VGGT-Long+Sim(3)-TurboReg}
& \textbf{320} & \textbf{705} & \textbf{149} & \textbf{756} & \textbf{109} & \textbf{34} & \textbf{433} & \textbf{173} & \textbf{158} & \textbf{619} & \textbf{225} & \textbf{161}
& \textbf{492} & \textbf{254} & \textbf{190} & \textbf{207} & \underline{837} & 920 & \textbf{577} & \textbf{273} & \textbf{769} & \textbf{403} \\
\rowcolor{gray!15}
\textbf{Pi-Long+Sim(3)-TurboReg}
& 1419 & 4899 & 440 & 2074 & 276 & 92 & 2354 & 664 & 595 & 2700 & 773 & 738
& 695 & 543 & 364 & \underline{312} & 1094 & 842 & 1145 & \underline{340} & 913 & 701 \\
\rowcolor{gray!15}
\textbf{DA3-Streaming+Sim(3)-TurboReg}
& \underline{529} & \underline{895} & \underline{186} & \underline{1314} & \underline{164} & \underline{48} & \underline{813} & \underline{255} & 346 & 1230 & \underline{329} & \underline{243}
& 641 & \underline{470} & 335 & 411 & 868 & \underline{795} & 1141 & 520 & \underline{771} & \underline{459} \\
\bottomrule
\end{tabular}
}
\end{table*}

%% file: conclusion.tex
\section{Conclusion}
\label{sec:conclusion}

We introduced Sim(3)-TurboReg, a robust and efficient Sim(3) registration solver built on a strict TSC criterion that suppresses both mismatching and coordinate-distortion outliers.
On controlled unknown-scale registration at 90\% outliers, the CPU variant matches RANSAC accuracy with $92.5\%$ lower median latency.
In DA3-Streaming on KITTI, the proposed backend improves mean ATE by $22.2\%$ and reduces mean end-to-end time by $16.6\%$.

% \section*{Acknowledgements}
% The authors thank the reviewers for their constructive feedback.
% This work was supported in part by [Funding sources to be added in the camera-ready version].

%% file: main.bib
@article{vggt_long,
  title={{VGGT-Long}: Chunk it, Loop it, Align it--Pushing {VGGT}'s Limits on Kilometer-scale Long {RGB} Sequences},
  author={Deng, Kai and Ti, Zexin and Xu, Jiawei and Yang, Jian and Xie, Jin},
  journal={arXiv:2507.16443},
  year={2025}
}

@article{laser,
  title={{LASER}: Layer-wise Scale Alignment for Training-Free Streaming {4D} Reconstruction},
  author={Ding, Tianye and Xie, Yiming and Liang, Yiqing and Chatterjee, Moitreya and Miraldo, Pedro and Jiang, Huaizu},
  journal={arXiv:2512.13680},
  year={2025}
}

@inproceedings{dust3r,
  title={{DUSt3R}: Geometric {3D} Vision Made Easy},
  author={Wang, Shuzhe and Leroy, Vincent and Cabon, Yohann and Chidlovskii, Boris and Revaud, Jerome},
  booktitle={Proc. IEEE/CVF Conf. Comput. Vis. Pattern Recognit. (CVPR)},
  year={2024}
}

@inproceedings{mast3r,
  title={Grounding Image Matching in {3D} with {MASt3R}},
  author={Leroy, Vincent and Cabon, Yohann and Revaud, Jerome},
  booktitle={Proc. Eur. Conf. Comput. Vis. (ECCV)},
  year={2024}
}

@inproceedings{cut3r,
  title={Continuous {3D} Perception Model with Persistent State},
  author={Wang, Qianqian and Zhang, Yifei and Holynski, Aleksander and Efros, Alexei A and Kanazawa, Angjoo},
  booktitle={Proc. IEEE/CVF Conf. Comput. Vis. Pattern Recognit. (CVPR)},
  year={2025}
}

@inproceedings{mast3r_slam,
  title={{MASt3R-SLAM}: Real-time Dense {SLAM} with {3D} Reconstruction Priors},
  author={Murai, Riku and Dexheimer, Eric and Davison, Andrew J},
  booktitle={Proc. IEEE/CVF Conf. Comput. Vis. Pattern Recognit. (CVPR)},
  year={2025}
}

@inproceedings{dpvo,
  title={Deep Patch Visual Odometry},
  author={Teed, Zachary and Lipson, Lahav and Deng, Jia},
  booktitle={Adv. Neural Inf. Process. Syst. (NeurIPS)},
  pages={39033--39051},
  year={2023}
}

@inproceedings{mac,
  title={{3D} Registration with Maximal Cliques},
  author={Zhang, Xiyu and Yang, Jiaqi and Zhang, Shikun and Zhang, Yanning},
  booktitle={Proc. IEEE/CVF Conf. Comput. Vis. Pattern Recognit. (CVPR)},
  year={2023}
}

@article{orbslam2,
  title={{ORB-SLAM2}: An Open-Source {SLAM} System for Monocular, Stereo, and {RGB-D} Cameras},
  author={Mur-Artal, Ra{\'u}l and Tard{\'o}s, Juan D.},
  journal={IEEE Trans. Robot.},
  volume={33},
  number={5},
  pages={1255--1262},
  year={2017}
}

@article{orbslam3,
  title={{ORB-SLAM3}: An Accurate Open-Source Library for Visual, Visual-Inertial, and Multimap {SLAM}},
  author={Campos, Carlos and Elvira, Richard and Rodr{\'i}guez, Juan J. G{\'o}mez and Montiel, Jos{\'e} M. M. and Tard{\'o}s, Juan D.},
  journal={IEEE Trans. Robot.},
  volume={37},
  number={6},
  pages={1874--1890},
  year={2021}
}

@inproceedings{droid_slam,
  title={{DROID-SLAM}: Deep Visual {SLAM} for Monocular, Stereo, and {RGB-D} Cameras},
  author={Teed, Zachary and Deng, Jia},
  booktitle={Adv. Neural Inf. Process. Syst. (NeurIPS)},
  pages={16558--16569},
  year={2021}
}

@inproceedings{dpvslam,
  title={Deep Patch Visual {SLAM}},
  author={Lipson, Lahav and Teed, Zachary and Deng, Jia},
  booktitle={Proc. Eur. Conf. Comput. Vis. (ECCV)},
  pages={424--440},
  year={2024}
}

@inproceedings{ldso,
  title={{LDSO}: Direct Sparse Odometry with Loop Closure},
  author={Gao, Xiang and Wang, Rui and Hu, Nikolaus and Zhu, Fenglong and Cremers, Daniel},
  booktitle={Proc. IEEE/RSJ Int. Conf. Intell. Robots Syst. (IROS)},
  year={2018}
}

@inproceedings{yang2024depth,
  title={{Depth Anything}: Unleashing the Power of Large-Scale Unlabeled Data},
  author={Yang, Lihe and Kang, Bingyi and Huang, Zilong and Xu, Xiaogang and Feng, Jianbing and Zhao, Hengshuang},
  booktitle={Proc. IEEE/CVF Conf. Comput. Vis. Pattern Recognit. (CVPR)},
  pages={10371--10381},
  year={2024}
}

@inproceedings{turboreg2025,
  title={{TurboReg}: {TurboClique} for Robust and Efficient Point Cloud Registration},
  author={Yan, Shaocheng and Shi, Pengcheng and Zhao, Zhenjun and Wang, Kaixin and Cao, Kuang and Wu, Ji and Li, Jiayuan},
  booktitle={Proc. IEEE/CVF Int. Conf. Comput. Vis. (ICCV)},
  year={2025}
}

@article{myronenko2010point,
  title={Point set registration: Coherent point drift},
  author={Myronenko, Andriy and Song, Xubo},
  journal={IEEE Trans. Pattern Anal. Mach. Intell.},
  volume={32},
  number={12},
  pages={2262--2275},
  year={2010},
  publisher={IEEE}
}

@article{jian2010robust,
  title={Robust point set registration using Gaussian mixture models},
  author={Jian, Bing and Vemuri, Baba C},
  journal={IEEE Trans. Pattern Anal. Mach. Intell.},
  volume={33},
  number={8},
  pages={1633--1645},
  year={2010},
  publisher={IEEE}
}

@article{yin2023metric3d,
  title={{Metric3D}: Towards Zero-shot Metric {3D} Prediction from A Single Image},
  author={Yin, Wei and Zhang, Chi and Chen, Hao and Cai, Zhipeng and Yu, Gang and Wang, Kaixuan and Chen, Xiaozhi},
  journal={arXiv:2307.10984},
  year={2023}
}

@inproceedings{vaswani2017attention,
  title={Attention is all you need},
  author={Vaswani, Ashish and Shazeer, Noam and Parmar, Niki and Uszkoreit, Jakob and Jones, Llion and Gomez, Aidan N and Kaiser, {\L}ukasz and Polosukhin, Illia},
  booktitle={Adv. Neural Inf. Process. Syst. (NeurIPS)},
  volume={30},
  year={2017}
}

@inproceedings{bertasius2021space,
  title={Is space-time attention all you need for video understanding?},
  author={Bertasius, Gedas and Wang, Heng and Torresani, Lorenzo},
  booktitle={Proc. Int. Conf. Mach. Learn. (ICML)},
  pages={813--824},
  year={2021}
}

@inproceedings{caesar2020nuscenes,
  title={{nuScenes}: A multimodal dataset for autonomous driving},
  author={Caesar, Holger and Bankiti, Varun and Lang, Alex H and Vora, Sourabh and Liong, Venice Erin and Xu, Qiangeng and Krishnan, Anush and Pan, Yu and Baldan, Giancarlo and Beijbom, Oscar},
  booktitle={Proc. IEEE/CVF Conf. Comput. Vis. Pattern Recognit. (CVPR)},
  pages={11621--11631},
  year={2020}
}

@article{cadena2016past,
  title={Past, present, and future of simultaneous localization and mapping: Toward the robust-perception age},
  author={Cadena, Cesar and Carlone, Luca and Carrillo, Henry and Latif, Yasir and Scaramuzza, Davide and Neira, Jos{\'e} and Reid, Ian and Leonard, John J},
  journal={IEEE Trans. Robot.},
  volume={32},
  number={6},
  pages={1309--1332},
  year={2016},
  publisher={IEEE}
}

@inproceedings{strasdat2010scale,
  title={Scale drift-aware large scale monocular {SLAM}},
  author={Strasdat, Hauke and Montiel, JMM and Davison, Andrew J},
  booktitle={Robot.: Sci. Syst. (RSS)},
  volume={2},
  number={3},
  pages={5},
  year={2010},
  organization={Zaragoza, Spain}
}

@inproceedings{zhang20233dmac,
  title={{3D MAC}: {3D} Maximal Cliques for Robust Point Cloud Registration},
  author={Zhang, Xinyi and Yang, Javen Qinfeng and Zhang, Shikun and Zhang, Yanning},
  booktitle={Proc. IEEE/CVF Conf. Comput. Vis. Pattern Recognit. (CVPR)},
  pages={3206--3215},
  year={2023}
}

@inproceedings{zhang2024fastmac,
  title={{FastMAC}: Stochastic Spectral Sampling of Correspondence Graph for Point Cloud Registration},
  author={Zhang, Xinyi and Yang, Javen Qinfeng and Zhang, Shikun and Zhang, Yanning},
  booktitle={Proc. IEEE/CVF Conf. Comput. Vis. Pattern Recognit. (CVPR)},
  year={2024}
}

@inproceedings{chen2022sc2,
  title={{SC$^2$-PCR}: A Second Order Spatial Compatibility for Efficient and Robust Point Cloud Registration},
  author={Chen, Zhi and Sun, Kun and Yang, Fan and Tao, Wenbing},
  booktitle={Proc. IEEE/CVF Conf. Comput. Vis. Pattern Recognit. (CVPR)},
  pages={13221--13231},
  year={2022}
}

@article{fischler1981random,
  title={Random Sample Consensus: A Paradigm for Model Fitting with Applications to Image Analysis and Automated Cartography},
  author={Fischler, Martin A and Bolles, Robert C},
  journal={Commun. ACM},
  volume={24},
  number={6},
  pages={381--395},
  year={1981}
}

@article{umeyama1991least,
  title={Least-Squares Estimation of Transformation Parameters Between Two Point Patterns},
  author={Umeyama, Shinji},
  journal={IEEE Trans. Pattern Anal. Mach. Intell.},
  volume={13},
  number={4},
  pages={376--380},
  year={1991}
}

@article{wang2025vggt,
  title={{VGGT}: Visual Geometry Grounded Transformer},
  author={Wang, Jianyuan and Chen, Minghao and Karaev, Nikita and Vedaldi, Andrea and Rupprecht, Christian and Novotny, David},
  journal={arXiv:2503.11651},
  year={2025}
}

@inproceedings{cabon2025must3r,
  title={{MUSt3R}: Multi-view Network for Stereo {3D} Reconstruction},
  author={Cabon, Yohann and Stoffl, Lucas and Antsfeld, Leonid and Csurka, Gabriela and Chidlovskii, Boris and Revaud, Jerome and Leroy, Vincent},
  booktitle={Proc. IEEE/CVF Conf. Comput. Vis. Pattern Recognit. (CVPR)},
  pages={1050--1060},
  year={2025}
}

@article{yang2020teaser,
  title={{TEASER}: Fast and Certifiable Point Cloud Registration},
  author={Yang, Heng and Shi, Jingnan and Carlone, Luca},
  journal={IEEE Trans. Robot.},
  volume={37},
  number={2},
  pages={314--333},
  year={2021},
  doi={10.1109/TRO.2020.3033695}
}

@article{HollandWelsch1977IRLS,
  title   = {Robust Regression Using Iteratively Reweighted Least-Squares},
  author  = {Holland, Paul W. and Welsch, Roy E.},
  journal={Commun. Stat., Theory Methods},
  volume  = {6},
  pages   = {813--827},
  year    = {1977},
  doi     = {10.1080/03610927708827533}
}

@article{depthanything3,
  title={{Depth Anything 3}: Recovering the Visual Space from Any Views},
  author={Haotong Lin and Sili Chen and Jun Hao Liew and Donny Y. Chen and Zhenyu Li and Guang Shi and Jiashi Feng and Bingyi Kang},
  journal={arXiv:2511.10647},
  year={2025}
}

@inproceedings{sturm12iros,
  title={A benchmark for the evaluation of {RGB-D} {SLAM} systems},
  author={Sturm, J{\"u}rgen and Engelhard, Nikolas and Endres, Felix and Burgard, Wolfram and Cremers, Daniel},
  booktitle={Proc. IEEE/RSJ Int. Conf. Intell. Robots Syst. (IROS)},
  pages={573--580},
  year={2012}
}

@inproceedings{geiger2012we,
  title={Are we ready for autonomous driving? The {KITTI} vision benchmark suite},
  author={Geiger, Andreas and Lenz, Philip and Urtasun, Raquel},
  booktitle={Proc. IEEE Conf. Comput. Vis. Pattern Recognit. (CVPR)},
  pages={3354--3361},
  year={2012}
}

@article{wang2025pi3,
  title     = {$\pi^3$: Scalable Permutation-Equivariant Visual Geometry Learning},
  author    = {Wang, Yifan and Zhou, Jianjun and Zhu, Haoyi and Chang, Wenzheng and Zhou, Yang and Li, Zizun and Chen, Junyi and Pang, Jiangmiao and Shen, Chunhua and He, Tong},
  journal   = {arXiv:2507.13347},
  year      = {2025}
}

@article{long3r,
  title     = {{LONG3R}: Long Sequence Streaming {3D} Reconstruction},
  author    = {Chen, Zhuoguang and Qin, Minghui and Yuan, Tianyuan and Liu, Zhe and Zhao, Hang},
  journal   = {arXiv:2507.18255},
  year      = {2025}
}

@article{longstream,
  title     = {{LongStream}: Long-Sequence Streaming Autoregressive Visual Geometry},
  author    = {Cheng, Chong and Chen, Xianda and Xie, Tao and Yin, Wei and Ren, Weiqiang and Zhang, Qian and Guo, Xiaoyuang and Wang, Hao},
  journal   = {arXiv:2602.13172},
  year      = {2026}
}

@article{smust3r,
  title     = {{S-MUSt3R}: Sliding Multi-view {3D} Reconstruction},
  author    = {Antsfeld, Leonid and Chidlovskii, Boris and Cabon, Yohann and Leroy, Vincent and Revaud, Jerome},
  journal   = {arXiv:2602.04517},
  year      = {2026}
}

@misc{da3_streaming,
  author       = {{ByteDance Seed}},
  title        = {{Depth Anything 3}: da3\_streaming (code)},
  howpublished = {\url{https://github.com/ByteDance-Seed/Depth-Anything-3/tree/main/da3_streaming}},
  year         = {2026},
  note         = {Accessed: Mar. 4, 2026}
}

@misc{stanford3dscanrep,
  author = {{Stanford University}},
  title = {The {Stanford} {3D} Scanning Repository},
  howpublished = {\url{https://graphics.stanford.edu/data/3Dscanrep/}},
  note = {Accessed: Sep. 16, 2026}
}

@inproceedings{zhou2016fast,
  author = {Zhou, Qian-Yi and Park, Jaesik and Koltun, Vladlen},
  title = {Fast Global Registration},
  booktitle={Proc. Eur. Conf. Comput. Vis. (ECCV)},
  year = {2016}
}
